%% file: main.tex
\documentclass[runningheads]{llncs}
\usepackage{tikz}
\usepackage{algorithmic}
\usepackage{graphicx}
\usepackage{amssymb}
\usepackage[cmex10]{amsmath}
\usepackage[linesnumbered,ruled,vlined]{algorithm2e}
\usepackage{multirow} 
\usepackage{multicol} 
\usepackage{arydshln}
\usepackage{graphicx}
\usepackage{epstopdf}
\usepackage{svg}
\usepackage{booktabs}
\usepackage{mdwmath}
\usepackage{mdwtab}
\usepackage{balance}
\definecolor{mylightgreen}{RGB}{144,238,144}
\definecolor{mylightred}{RGB}{255,66,53}
\definecolor{mylightblue}{RGB}{218,227,245} 
\definecolor{darkred}{rgb}{0.70, 0, 0}
\definecolor{darkgreen}{rgb}{0, 0.55, 0}
\definecolor{darkblue}{rgb}{0, 0.0, 0.78}
\definecolor{darkpurple}{rgb}{0.53, 0, 0.50}
\definecolor{purple}{rgb}{0.57, 0.55, 0.78}
\definecolor{iryellow}{rgb}{0.66, 0.82, 0.56}
\definecolor{trolleygrey}{rgb}{0.5, 0.5, 0.49}
\definecolor{tropicalrainforest}{rgb}{1.0, 0.91, 0.7}
\definecolor{glaucous}{rgb}{0.38, 0.51, 0.71}
\definecolor{cardinal}{rgb}{0.7, 0.33, 0.3}
\definecolor{palegreen}{rgb}{0.61, 0.7, 0.35}
\definecolor{pink}{rgb}{0.97, 0.78, 0.65}
\definecolor{orange}{rgb}{0.96, 0.69, 0.51}
\definecolor{purple}{rgb}{0.57, 0.55, 0.78}

\begin{document}

% \title{ProKG-Dial: A Progressive LLM-Driven Approach to Building Knowledge-Intensive Multi-Turn Dialogue Datasets with Domain Knowledge Graphs
% }

\title{ProKG-Dial: Progressive Multi-Turn Dialogue Construction with Domain Knowledge Graphs
}

\author{Yuanyuan Liang\inst{1} \and
Xiaoman Wang \inst{1}  \and 
Tingyu Xie\inst{2} \and
 Lei Pan\inst{3} 
}
\authorrunning{F. Author et al.}
\institute{East China Normal University, China\\
\email{\{leonyuany,xmwang\}@stu.ecnu.edu.cn}\\ 
\and
Zhejiang University, China\\
\email{tingyuxie@zju.edu.cn} \\
\and
AISpeech, China\\
\email{lei.pan@aispeech.com}\\
}
\maketitle

\begin{abstract}
Current large language models (LLMs) excel at general NLP tasks but often lack domain-specific precision in professional settings. Building a high-quality, domain-specific multi-turn dialogue dataset is essential for developing specialized conversational systems. However, existing methods—such as manual annotation, simulated human-LLM interactions, and role-based LLM dialogues—are resource-intensive or suffer from limitations in dialogue quality and domain coverage.
To address these challenges, we introduce \textbf{ProKG-Dial}, a progressive framework for constructing knowledge-intensive multi-turn dialogue datasets using domain-specific knowledge graphs (KGs). ProKG-Dial leverages the structured nature of KGs to encode complex domain knowledge and relationships, providing a solid foundation for generating meaningful and coherent dialogues.
Specifically, ProKG-Dial begins by applying community detection to partition the KG into semantically cohesive subgraphs. For each subgraph, the framework incrementally generates a series of questions and answers centered around a target entity, ensuring relevance and coverage. A rigorous filtering step is employed to maintain high dialogue quality.
We validate ProKG-Dial on a medical knowledge graph by evaluating the generated dialogues in terms of diversity, semantic coherence, and entity coverage. Furthermore, we fine-tune a base LLM on the resulting dataset and benchmark it against several baselines. Both automatic metrics and human evaluations demonstrate that ProKG-Dial substantially improves dialogue quality and domain-specific performance, highlighting its effectiveness and practical utility.

\keywords{
Multi-turn dialogue \and domain knowledge graph \and large language model}
\end{abstract}

\input{intro}
\input{related}

\input{method}
\input{expr}

\section{Conclusion}
In this paper, we present a framework for constructing domain-specific, knowledge-intensive multi-turn dialogue datasets starting from a domain KG. By integrating community partitioning and generating questions and answers based on entities within subgraphs, our approach ensures topic consistency and alignment. We applied this framework to create a medical domain dialogue dataset, and both automatic and human evaluations demonstrate its high quality. Our method offers a scalable solution for enhancing dialogue systems in specialized domains, with potential for expansion to other fields in the future.

\section{Limitations}
While ProKG-Dial demonstrates strong performance in generating domain-specific multi-turn dialogues, it also has limitations. The reliance on existing KGs means the quality and completeness of the KG directly affect the generated dialogue quality. Moreover, the semantic filtering process may mistakenly remove valid variations if subgraphs are semantically similar but contextually different. Future work could explore more nuanced semantic clustering and integrate external knowledge sources to mitigate KG incompleteness.

\bibliographystyle{splncs04}
\bibliography{sample-base}
\end{document}

%% file: intro.tex
% \begin{figure}
% \fbox{\begin{minipage}{0.47\textwidth}
% \textcolor{blue}{\textbf{A}}: I'm \textcolor{purple}{\underline{worried}} about something.\\
% \textcolor{red}{\textbf{B}}: What's that?\\
% \textcolor{blue}{\textbf{A}}: Well, I have to drive to school for a meeting this morning, and I'm going to end up getting stuck in rush-hour traffic.\\
% \textcolor{red}{\textbf{B}}: That's \textcolor{purple}{annoying}, but nothing to worry about. \emph{Just breathe deeply when you feel yourself getting upset}.\\
% \textcolor{blue}{\textbf{A}}: Ok, I'll try that.\\
% \textcolor{red}{\textbf{B}}: Is there anything else \textcolor{purple}{\underline{bothering}} you?\\
% \textcolor{blue}{\textbf{A}}: Just one more thing. A school called me this morning to see if I could teach a few classes this weekend and I don't know what to do.\\
% \textcolor{red}{\textbf{B}}: Do you have any other plans this weekend?\\
% \textcolor{blue}{\textbf{A}}: I'm supposed to work on a paper that'd due on Monday.\\
% \textcolor{red}{\textbf{B}}: \emph{Try not to take on more than you can handle}.\\
% \textcolor{blue}{\textbf{A}}: You're right. I probably should just work on my paper. \textcolor{purple}{\underline{Thanks}}!
% \end{minipage}
% }
% \caption{An example of multi-turn dialogue in the medical field.}
% \label{fig:example}
% \end{figure}
\vspace{-1.5em}
\begin{figure}[ht]
\fbox{\begin{minipage}{0.99\textwidth}
\textcolor{green}{\textbf{A}}: I'm evaluating my \textcolor{red}{investment portfolio}, but recent \textcolor{red}{interest rate hikes} have me concerned. How might this affect my \textcolor{red}{bond holdings}?\\
\textcolor{blue}{\textbf{B}}: Rising interest rates typically cause \textcolor{red}{bond prices} to fall, especially for \textcolor{red}{long-duration bonds}. Have you considered shifting towards \textcolor{red}{shorter maturities} to reduce \textcolor{red}{interest rate risk}?\\
\textcolor{green}{\textbf{A}}: That makes sense. What about \textcolor{red}{equity investments}—should I be worried about volatility in the \textcolor{red}{financial sector}?\\
\textcolor{blue}{\textbf{B}}: Volatility can be higher in that sector during rate hikes. Diversifying into \textcolor{red}{defensive stocks} or \textcolor{red}{sectors less sensitive} to rate changes, like \textcolor{red}{utilities} or \textcolor{red}{consumer staples}, might help balance your risk.\\
\textcolor{green}{\textbf{A}}: Understood. Also, my advisor suggested increasing my exposure to \textcolor{red}{emerging market equities}. Given the current \textcolor{red}{geopolitical tensions} and \textcolor{red}{currency fluctuations}, is this advisable?\\
\textcolor{blue}{\textbf{B}}: Emerging markets offer growth potential but come with higher \textcolor{red}{political} and \textcolor{red}{currency risks}. It’s important to evaluate your \textcolor{red}{risk tolerance} and consider hedging strategies or limiting allocation accordingly.\\
\textcolor{green}{\textbf{A}}: Thanks, that helps clarify. Lastly, should I adjust my \textcolor{red}{asset allocation} to account for \textcolor{red}{inflation expectations}?\\
\textcolor{blue}{\textbf{B}}: Yes, inflation can erode real returns. Incorporating \textcolor{red}{inflation-protected securities} or \textcolor{red}{commodities} may help preserve purchasing power. Regularly reviewing your portfolio in light of macroeconomic trends is key.\\
\textcolor{green}{\textbf{A}}: Got it. I’ll discuss these adjustments with my advisor. Thanks for your insights!
\end{minipage}
}
\caption{An example of a domain-specific multi-turn dialogue in finance, where green represents the user's questions, blue the responses, and red highlights key domain terms.}
\label{fig:example}
\vspace{-1.5em}
\end{figure}

\section{Introduction}
Multi-turn dialogue is an interactive process that involves multiple rounds, where the user asks questions and receives responses from the model in each round. As shown in ~\ref{fig:example}, the interaction between the user and the model can focus on a specific domain, such as the financial field, with multiple rounds of question-and-answer exchanges. In each round, the model provides targeted responses and adjusts its strategy based on the preceding information.
Current LLMs possess powerful general-purpose natural language processing capabilities, enabling them to efficiently perform tasks such as text generation, question answering, and language translation~\cite{lee2024llm2llm,zhao2023survey}. However, when handling professional conversations in specific domains, they may encounter knowledge gaps or generate responses that lack precision~\cite{li2023large,pan2024unifying}. This is primarily because these models are typically trained on extensive general-purpose corpora, with insufficient coverage of in-depth specialized knowledge in specific fields. For instance, conversations in domains such as medicine, law, or finance often involve technical terminology, complex reasoning, and domain-specific background knowledge~\cite{wu2023bloomberggpt}. General-purpose language models may struggle to accurately understand users' intentions or generate responses that meet industry standards.

Moreover, domain-specific dialogues often exhibit continuity and contextual dependency, requiring the model to retain and utilize prior information to respond appropriately to subsequent questions~\cite{byambasuren2019preliminary}. However, general models may lose context in multi-turn conversations, leading to responses that lack coherence or logical consistency. Therefore, to build a conversational system that truly meets the needs of specialized domains, it is essential to collect and construct multi-turn dialogue datasets specific to the target domain.

Currently, three common methods are used to construct multi-turn dialogue datasets: first, manual annotation; second, dialogues between humans and LLMs in simulated scenarios; and third, assigning different roles to multiple LLMs to engage in conversations on specific topics~\cite{ding2023enhancing}. 
The first two methods require domain experts, which not only demands substantial human and material resources but also incurs significant time costs. While these approaches ensure a certain level of professionalism, their applicability and efficiency are limited by their dependence on expert knowledge. The third method, while not reliant on domain experts, faces challenges in effectively controlling dialogue quality.
Without sufficient supervision mechanisms, it becomes difficult to guarantee the accuracy and professionalism of the content. Furthermore, all three methods share a common limitation: they struggle to ensure that the generated data comprehensively covers the knowledge of a specific domain, which results in gaps in both the breadth and depth of knowledge within the dialogue data. Additionally, the agent's responses often become overly long and complex, increasing the cognitive load on the user~\cite{li2017dailydialog}.

In some specialized fields, KGs are commonly used to model complex data relationships, integrate information, and perform knowledge reasoning, such as in healthcare, finance, recommendation systems, and search engines~\cite{pan2024unifying,chen2020review}. These fields often involve large amounts of specialized knowledge and complex structured information, which require the powerful capabilities of KGs to effectively organize and reason about information~\cite{zou2020survey,vuth-etal-2024-kgast}. 
KGs serve as a surrogate for expert knowledge by encoding domain expertise into structured triples, thus replacing part of the manual effort typically required for dialogue design. 

To address these limitations, we leverage domain-specific KGs as structured alternatives to manual expertise. KGs encode rich domain knowledge through entities and relationships, enabling automatic guidance for dialogue generation and reducing dependence on human effort. By modeling the semantics of professional fields, KGs help large language models produce accurate, coherent, and knowledge-grounded multi-turn dialogues while significantly lowering human and material costs. Building on this, we propose a framework for constructing knowledge-intensive multi-turn dialogue datasets driven by domain-specific KGs, enhancing professionalism, coverage, and reasoning quality to improve domain-specific conversational systems.

The process begins with community partitioning of the domain KG to identify subgraphs of tightly connected entities and relations~\cite{mehta2024using,christensen2024comparing}. This step divides the KG into smaller, cohesive regions, facilitating focused analysis on relevant areas. Community partitioning not only uncovers potential domain features but also reveals latent relationships, providing a precise foundation for downstream analysis and question generation. Within each subgraph, a target entity is selected as a starting point, and an incremental approach generates a series of closely related questions and answers based on that entity and its connections. This stepwise strategy ensures dialogues remain relevant and diverse while avoiding irrelevant or redundant content, thereby improving overall data quality and utility.

Finally, the generated multi-turn dialogue data undergo rigorous filtering to remove low-quality samples, retaining those with high information density and semantic accuracy. Through this quality control, the resulting domain-specific dialogue dataset is well-structured and richly representative of the domain knowledge. These high-quality datasets provide strong support for domain-specific NLP tasks, enhancing model performance and effectiveness.

We summarize our contributions as follows:
\begin{itemize}
    \item 
     We propose a framework for gradually constructing domain-specific, knowledge-intensive multi-turn dialogue data starting from the domain KG. 
     % ~\krcomment{Maybe clarify the methods of generating NL-GQL pairs.}
    \item 
     Our approach incorporates community partitioning in the KG and progressively generates questions and answers based on entities in subgraphs, ensuring topic consistency and alignment between questions and answers.
    \item
     Using this method, we created a medical domain multi-turn dialogue dataset. Analysis shows the effectiveness of our approach, and fine-tuning open-source LLMs with this dataset improves dialogue system performance, as confirmed by both automatic and human evaluations.
\end{itemize}

%% file: related.tex
\section{Related Work}

\subsection{KG-Driven Synthetic Data Generation using LLMs. }
% \noindent \textbf{KG-Driven Synthetic Data Generation using LLMs. }
In the rapidly evolving field of deep learning, the challenge of missing data has become increasingly prominent. The emergence of LLMs presents a data-centric solution to address the limitations of real-world data by generating synthetic data~\cite{long2024llms,zhou2024survey}. LLMs are capable of producing vast volumes of diverse, high-quality synthetic data tailored to specific domain needs, effectively mitigating data scarcity. Their ability to replicate the statistical properties, structures, and patterns of real datasets makes them highly adaptable across various industries, including healthcare and finance~\cite{tang2023does,li2023large,wu2023bloomberggpt}. By generating data that closely mirrors real-world scenarios, LLMs not only augment existing datasets but also enable model training under diverse conditions that would otherwise be difficult to simulate. Furthermore, their deep contextual understanding ensures that the synthetic data generated is both relevant and meaningful, enhancing the overall quality and effectiveness of model training~\cite{lee2024llm2llm,abdullin2024synthetic}.

In addition to the capabilities of LLMs, KGs offer a natural extension for generating synthetic data~\cite{chen2020review,pan2024unifying}. KGs encode structured, interconnected information about entities, relationships, and attributes, making them ideal for representing complex real-world scenarios~\cite{meyer2023llm}. By capturing semantic relationships and hierarchical structures, KGs provide a rich framework for contextualizing data and generating more coherent and realistic synthetic examples~\cite{agarwal2020knowledge,kang2024artificial}. Their structured nature enables the representation of both explicit and implicit knowledge, offering a distinct advantage in creating diverse data that reflects the complexities and nuances found in real-world systems~\cite{pan2024unifying,zou2020survey}. This makes KGs an excellent tool for generating synthetic data that adheres to domain-specific semantics while preserving critical relationships between entities. By integrating KGs into the data generation process, LLMs enhance their understanding of entities, relationships, and domain-specific structures, leading to the creation of complex synthetic datasets that better align with real-world knowledge\cite{zhou2024survey}. 

The synergy between LLMs and KGs provides a powerful approach to synthetic data generation. LLMs can leverage the knowledge embedded in KGs to generate contextually accurate and semantically rich synthetic data~\cite{pan2024unifying}. 
For example, \cite{xu2023knowledge} utilizes KGs and LLMs to extract domain topics for more context-driven prompting, thereby enhancing data fidelity and complexity. 
The work \cite{agarwal2020knowledge} explores converting the English Wikidata KG into natural text, integrating structured knowledge with natural language to improve factual accuracy and performance on knowledge-intensive tasks. 
The paper \cite{kumichev2024medsyn} introduces MedSyn, a framework that combines LLMs with a Medical KG to generate synthetic medical texts, aiming to improve the quality and diversity of medical data. 
Research presented in \cite{vuth-etal-2024-kgast} introduces KGAST, a novel synthetic data generation framework that uses KGs and LLMs to automatically generate annotated synthetic data for Information Extraction tasks. 
The paper \cite{construction-paired-knowledge-graph} discusses constructing improved datasets using heuristics to enhance the equivalence between KGs and text, utilizing LLMs to generate synthetic data that optimizes performance on cyclic generation tasks.

\subsection{Multi-Turn Dialogue Datasets Generation.}
% \noindent \textbf{Multi-Turn Dialogue Datasets Generation. }
The generation of multi-turn dialogue datasets has become increasingly vital with the advancement of LLMs. Traditional methods of data collection often involve manual annotation, which is both time-consuming and resource-intensive\cite{yi2024survey}.

To address this, researchers have turned to LLMs to automate the creation of high-quality dialogue datasets. For instance, the study ~\cite{stacey2024lucid} introduces a modularized and highly automated LLM-driven data generation system that produces realistic, diverse, and challenging dialogues datasets. 
Another notable contribution is the ~\cite{yang2023refgpt}, which proposes a method to generate truthful and customized dialogues by leveraging LLMs.
In the realm of multi-modal dialogues, the paper ~\cite{feng-etal-2023-mmdialog} offers a comprehensive dataset comprising 1.08 million real-world dialogues with 1.53 million unique images across 4,184 topics.
The paper~\cite{qiu2024smile} introduces a method that utilizes ChatGPT to transform single-turn dialogues into multi-turn conversations, thereby creating a diverse and high-quality dataset named SmileChat to enhance mental health chatbot development.
The paper \cite{yang2023zhongjing} constructs the CMtMedQA dataset, comprising 70,000 authentic doctor-patient dialogues, which significantly enhances the model's ability to handle complex dialogues and initiate proactive inquiries.
Furthermore, the work ~\cite{wen2023re3dial} addresses the scarcity of long-turn dialogues in existing pre-training corpora. By reorganizing short-turn dialogues, this method constructs long-turn dialogues, thereby improving the model's ability to utilize long-range context and generate more coherent and informative responses. 

% Unlike existing methods, our approach uses domain-specific KGs and community partitioning to partition the graph into subgraphs, generating dialogues that focus on closely related entities and relationships. By iteratively expanding from specific entities, we ensure diverse and high-quality dialogues with comprehensive domain coverage. This method enhances the accuracy and relevance of dialogues, addressing the limitations of current approaches in handling specialized knowledge.

Unlike existing methods, our approach uses domain-specific KGs and community partitioning to partition the graph into subgraphs, generating dialogues that focus on closely related entities and relationships. By iteratively expanding from specific entities, we ensure diverse and high-quality dialogues with comprehensive domain coverage. This method enhances the accuracy and relevance of dialogues, addressing the limitations of current approaches in handling specialized knowledge. \textbf{In contrast to existing resource-heavy approaches, our method exploits KGs to automatically guide dialogue generation, minimizing the need for costly domain experts.}

%% file: method.tex
\section{Preliminaries}

\noindent \textbf{KG.}
A KG is a graph structure that represents entities and their relationships through nodes and edges. Nodes typically represent entities (such as people, places, things, or concepts), while edges represent the various relationships between entities (such as "belongs to," "located in," "owns," etc.). A subgraph of a KG refers to a subset extracted from the entire graph, containing specific nodes and the relationships between them. The KG and subgraph can be formulated as follows:

\begin{equation} \mathcal{KG} = {(e_1, r, e_2) \mid e_1, e_2 \in V, r \in E} \end{equation}
\begin{equation} \mathcal{KG}_{sub} = {(e_1', r', e_2') \mid e_1', e_2' \in V', r' \in E'} \end{equation}

Here, $E$ and $V$ denote the sets of vertices and edges, respectively, while $E'$ and $V'$ are subsets of $E$ and $V$.

\noindent \textbf{Multi-turn Dialogue.}
Multi-turn dialogue refers to a process in which the system and the user engage in multiple interactions, with each turn building upon the previous dialogue content. Unlike single-turn dialogue, multi-turn dialogue requires the system to remember contextual information and generate appropriate responses based on the historical dialogue state. This can be formulated as follows:
\begin{equation} 
s_t = (u_1, s_1), (u_2, s_2), \dots, (u_{t-1}, s_{t-1}), (u_t) 
\end{equation}

Here, \( u_t \) represents the question in the \( t \)-th round, and \( s_t \) represents the response in turn \( t \).

\begin{figure*}[ht]
\centering
\includegraphics[width=0.98\textwidth]{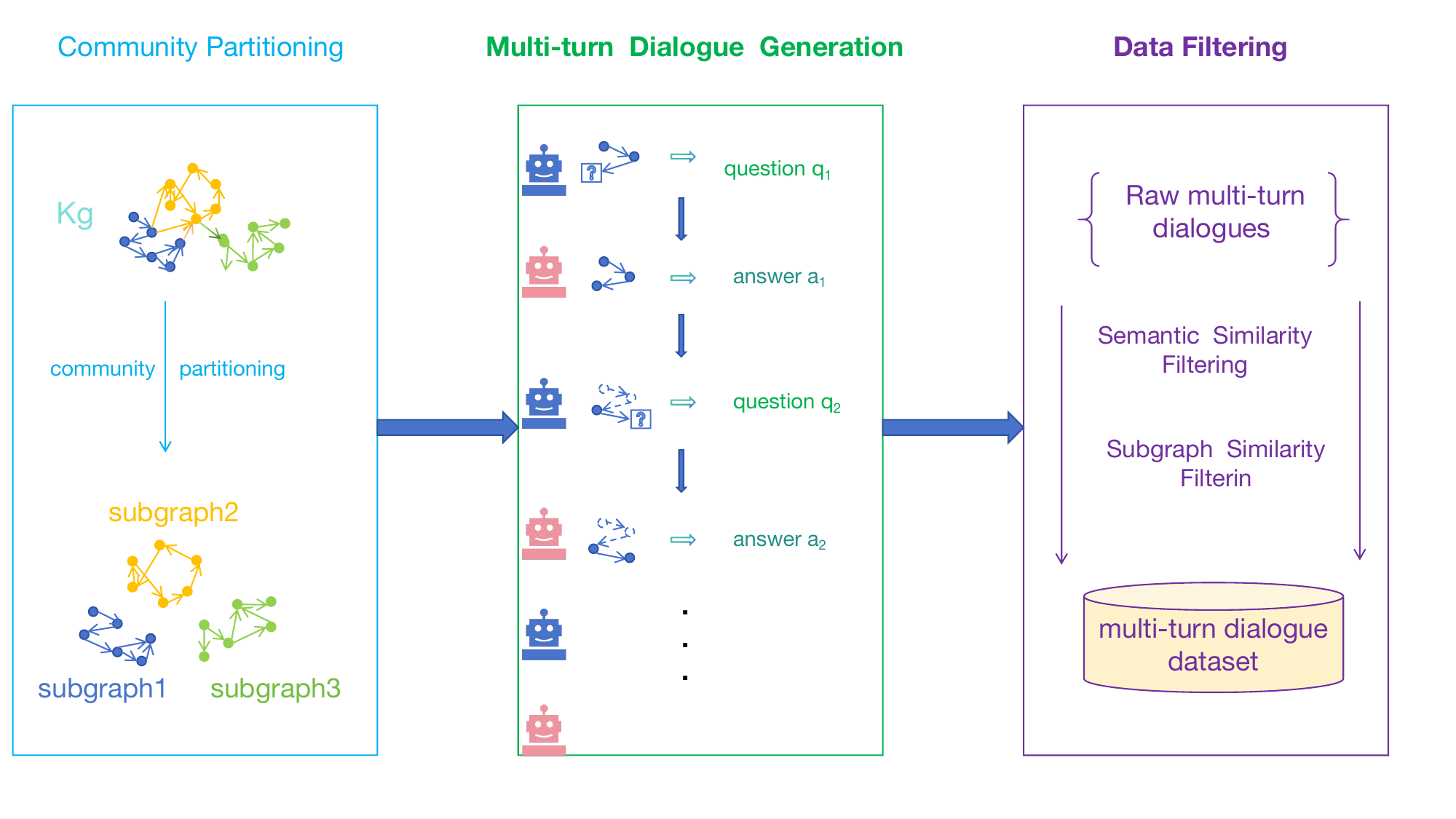}
\caption{
Our ProKG-Dial method consists of three components. On the far left, it involves community partitioning of the KG. The middle component features the collaboration between the question generation agent and the response agent, which progressively guides the answer generation for multi-turn dialogue based on the entities from the subgraph. The final component focuses on data filtering.
}
\label{fig:Overview}
\vspace{-1.5em}
\end{figure*}

\section{Method}
In this section, we will provide a detailed explanation of our ProKG-Dial method. As shown in \ref{fig:Overview}, our method consists of three components: community partitioning, multi-turn dialogue generation, and data filtering.

\subsection{Graph Embedding-Enhanced Community Partitioning}
% ----------------------------社区划分伪代码-------------------
\begin{algorithm}[htbp]
\caption{Graph Community Partitioning with GraphSAGE Embeddings}
\label{algo_comm} 
\KwIn{$\mathcal{KG} = {(e_1, r, e_2) \mid e_1, e_2 \in V, r \in E}$}
\KwIn{$\mathcal{E} = \{e_1, e_2, \dots, e_n\}$: Node embeddings from GraphSAGE}
\KwOut{$C = \{C_1, C_2, \dots, C_k\}$: Communities in graph $G$}
\textbf{Step 1: Node Embedding Initialization} \\
\hspace{0.2cm} $\mathcal{E} \leftarrow$ GraphSAGE($G$) \\
\hspace{0.2cm} Normalize $\mathcal{E}$ \\
\hspace{0.2cm} Assign initial communities: $C_i = \{v_i\}$ for $v_i \in V$
\textbf{Step 2: Community Merging Based on Embeddings} \\
\For{each edge $(u, v) \in E$}{
    \hspace{0.2cm} $Sim(e_u, e_v) \leftarrow \text{similarity}(e_u, e_v)$ \\
    \hspace{0.2cm} \If{$Sim(e_u, e_v) > \theta$}{
        \hspace{0.4cm} Propose merge: $C_u \cup C_v$
    }
}
\textbf{Step 3: Louvain Optimization} \\
\hspace{0.2cm} $Q \leftarrow$ Modularity of current partition \\
\For{each node $v_i \in V$}{
    \hspace{0.2cm} Try move $v_i$ to $C_j \neq C_i$ that maximizes $Q$
} 
\While{$Q$ improves}{
    \hspace{0.2cm} Repeat above until no significant change in $Q$
}
\textbf{Step 4: Return Communities}
\hspace{0.2cm} Output final community structure $C = \{C_1, C_2, \dots, C_k\}$ \\

\end{algorithm}

In our dialogue generation task, the primary objective of community partitioning is to capture and optimize the relationships between nodes using graph embedding techniques, thereby improving the quality of multi-turn dialogue generation. By partitioning the graph, we can identify highly interconnected node groups and treat these as interacting entity clusters. This partitioning facilitates the generation of contextually relevant dialogue, as user needs and system responses in multi-turn dialogues are often closely tied to specific entity groups. In the Louvain algorithm, we incorporate the graph embedding information of nodes into a weighted optimization process, ensuring that the community partitioning not only considers node connectivity but also effectively leverages the high-dimensional feature representations of the nodes, thus enhancing the quality of the partitioning. As shown in Algorithm \ref{algo_comm}, our algorithm consists of four steps.

\noindent \textbf{Step 1: Node Embedding Initialization.}
We first generate node embeddings using GraphSAGE, which aggregates information from a node's local neighborhood. The embeddings $\mathcal{E} = \{e_1, e_2, \dots, e_n\}$ are then normalized. Initially, each node $v_i$ is assigned to its own community: $C_i = \{v_i\}$.

\noindent \textbf{Step 2: Community Merging.}
For each edge $(u, v) \in E$, we compute the similarity $Sim(e_u, e_v)$. If $Sim(e_u, e_v)$ exceeds a threshold $\theta$, we propose merging the communities of nodes $u$ and $v$: $C_u \cup C_v$.

\noindent \textbf{Step 3: Louvain Optimization.} 
We compute the modularity $Q$ of the current partition. For each node $v_i$, we attempt to move it to a different community $C_j$ that maximizes $Q$. This process continues until $Q$ no longer improves.

\noindent \textbf{Step 4: Return Communities.} 
The final community structure is returned as $C = \{C_1, C_2, \dots, C_k\}$, where each $C_i$ contains the nodes assigned to it.

Through this module, we obtain some tightly connected subgraphs, preparing for the next step of multi-turn dialogue generation.

\subsection{Multi-turn Dialogue Generation}

\begin{algorithm}[ht!]
\caption{ARGW Algorithm}
\label{argw_algo}
\begin{algorithmic}[1]
\STATE \textbf{Input:} KG $G$, Initial entity $e(i)$, Number of steps $T$
\STATE \textbf{Output:} Set of generated questions and answers

\STATE \textbf{Initialize:}
\STATE $current\_entity \gets e(i)$
\STATE $walk\_history \gets []$
\STATE $questions \gets []$
\STATE $answers \gets []$

\FOR{$t = 1$ \textbf{to} $T$}
    \STATE \textbf{Step 1:} Select relations $R_{current\_entity}$ of $current\_entity$
    \STATE \textbf{Step 2:} Dynamically adjust relation weights based on semantic importance and graph structure
    \STATE \textbf{Step 3:} Select next entity $e(i+1)$ by walking along relation $r(i)$ with highest weight
    \STATE \textbf{Step 4:} Update $walk\_history$ with $(current\_entity, r(i), e(i+1))$
    
    \STATE \textbf{Step 5:} Based on $walk\_history$, generate a question related to $e(i+1)$
    \STATE \textbf{Step 6:} Use domain knowledge or graph to generate the corresponding answer for $e(i+1)$
    \STATE Add the generated question to $questions$
    \STATE Add the corresponding answer to $answers$
    
    \STATE \textbf{Step 7:} Check for redundant paths and prune if necessary
    \STATE \textbf{Step 8:} If $e(i+1)$ has similar entities $e'(i+1)$, expand the walk with $e'(i+1)$ and its relation $r(i+1)$
    \STATE $current\_entity \gets e(i+1)$
\ENDFOR

\STATE \textbf{Return:} Set of generated questions and answers $(questions, answers)$
\end{algorithmic}
\end{algorithm}

To ensure that multi-turn dialogues cover sufficient entity knowledge, stay on topic, and maintain diversity, we designed the Adaptive Relationship-guided Graph Walk (ARGW) algorithm. ARGW guides entity expansion within a graph, generating questions closely tied to entities and relationships. It combines structural and semantic information, utilizing adaptive relation selection, entity similarity, and redundancy elimination to avoid irrelevant questions while ensuring diversity and semantic accuracy. The algorithm steps are outlined in Algorithm~\ref{argw_algo}. 

The ARGW Algorithm begins by initializing with an initial entity \( e(i) \), a specified number of steps \( T \), and empty lists for the walk history, questions, and answers. For each step, the algorithm first identifies all possible relations connected to the current entity and dynamically adjusts their weights based on semantic importance and the graph's structure. It then selects the next entity \( e(i+1) \) by walking along the relation with the highest adjusted weight, and updates the walk history with the current entity, relation, and newly selected entity. Using the walk history, the algorithm generates a question related to \( e(i+1) \), followed by generating the corresponding answer based on domain knowledge or the graph. To ensure the dialogue remains focused, the algorithm checks for redundant paths and prunes them if necessary, before continuing the expansion.

The graph expansion algorithm for the entity knowledge points involved in the dialogue was introduced earlier. Next, we will detail the specific method for generating multi-turn dialogue data. We assign distinct roles to two LLMs: a Question Generator and an Answer Generator. The Question Generator formulates the next inquiry based on dialogue history and graph expansion, while the Answer Generator responds using the same context. This separation simulates human-agent interaction more realistically and allows for independent control and optimization of each dialogue turn. In each round of dialogue, the question model generates the next question based on the current dialogue history and the expansion path for the next question determined by the ARGW algorithm. Similarly, the answer model generates the corresponding answer based on the current dialogue history and the expansion path for the next answer, also guided by the ARGW algorithm. The prompt structure is shown in Figure~\ref{fig:dia_generate}.

At the end of each round, both models update their respective dialogue histories, which then serve as the context for the next round. This iterative interaction creates a feedback loop, ensuring that the entire dialogue process remains logically coherent and semantically rich. Through this process, we guarantee that the generated multi-turn dialogues not only cover the entities and relationships within the KG but also preserve the diversity, information density, and semantic accuracy of the conversation.

\begin{figure}[ht]
\centering
\resizebox{0.99\textwidth}{!}{
  \begin{tikzpicture}
    \node[draw, rectangle, inner sep=2mm, fill=mylightblue] (rect) {
      \begin{minipage}{\linewidth}
      \textbf{Instruction:}\\
You are an expert in the NLP field. I would appreciate your assistance with a data generation task. Based on the provided historical dialogue information and the subgraph query path (or answer entity) for the next round of dialogue, please generate the corresponding natural language question (or natural language answer). The generated questions (or answers) should be concise, clear, and no longer than 30 characters.
\\

\textbf{Output Format: }\\
Directly output the generated natural language question (natural language answer).\\

\textbf{Here is the dialogue history:}\\
\{DIALOGUE HISTORY\}\\
\textbf{Here is the subgraph query path (the entities related to the question and answer):}\\
\{SUBGRAPH QUERY PATH OR ANSWER ENTITY \}\\
\textbf{Output:}\\
      \end{minipage}
    };
  \end{tikzpicture}
}
\caption{Prompt for generating dialogue questions or answers.}
\label{fig:dia_generate}
\vspace{-1.5em}
\end{figure}

\subsection{Data Filtering}
After generating the original multi-turn dialogue data, we need to filter out redundant and highly similar data to ensure that the final dataset is both diverse and meaningful. To achieve this, we have designed a dual filtering method that combines semantic embedding similarity filtering with subgraph-based multi-turn dialogue filtering.

\begin{table*}[ht]

\centering
\caption{\label{static_result}
Basic Statistics of The Dataset.}
\resizebox{0.96\textwidth}{!}{
\tabcolsep=1.1em
\renewcommand\arraystretch{0.8}
\begin{tabular}{lllll}
\toprule
                                   & train  & dev   & test  & total  \\ \midrule
Total Dialogues                    & 4800   & 800   & 1600  & 7200   \\ \midrule
Average Speaker Turns Per Dialogue & 8.2    & 8.1   & 8.4   & 8.2    \\ \midrule
Average Tokens Per Dialogue        & 126.6  & 137.2 & 141.7 & 131.1  \\ \midrule
Total Key Entities                 & 232800 & 36560 & 80480 & 349840 \\ \midrule
Average Key Entities Per Dialogue  & 48.5   & 45.7  & 50.3  & 48.6   \\ \bottomrule
\end{tabular}
}

\vspace{-1.5em}
\end{table*}

\noindent \textbf{Semantic  Similarity Filtering}  
To identify and remove redundant dialogue data, we employ pre-trained language models to generate semantic embeddings for each piece of multi-turn dialogue data, resulting in high-dimensional vector representations. These vectors effectively capture the semantic content of the dialogues and form the basis for subsequent similarity calculations. We then compute the similarity between dialogues using cosine similarity or other distance metrics. If the similarity exceeds a predefined threshold, the dialogues are flagged as redundant, and one of them is removed.

\noindent \textbf{Subgraph  Similarity Filtering}  
Multi-turn dialogues often involve specific entities and relationships that form distinct structures in the KG. We extract the relevant subgraphs from the KG and analyze the nodes and relationships within them. Each dialogue is associated with a specific subgraph, and multiple dialogues may share similar or identical subgraphs. To detect duplicate or highly similar subgraphs, we calculate the overlap rate between them. If certain dialogues involve highly similar subgraphs, those dialogues are considered redundant and filtered out. We define subgraph similarity using the Jaccard Index over the sets of triples:
\[
\text{Sim}(G_1, G_2) = \frac{|E_{G_1} \cap E_{G_2}|}{|E_{G_1} \cup E_{G_2}|}
\]
where $E_{G_1}$ and $E_{G_2}$ represent the edge sets of two subgraphs.
If $\text{Sim}(G_1, G_2) > \tau$, one of them is considered redundant and removed.

After filtering, we obtain a more diverse and high-quality multi-turn dialogue dataset. This dataset not only eliminates redundancy and repetition but also ensures significant semantic variation in each dialogue, enabling the dialogue generation model to learn from a broader range of dialogue scenarios and rich contextual information.

%% file: expr.tex
\section{Experiments and Analysis}
\subsection{Experimental Setup}
\noindent \textbf{KG } 
We use CMeKG~\cite{byambasuren2019preliminary} as the foundation of our KG. Built with natural language processing, text mining, and human-machine collaboration, CMeKG incorporates standards like ICD, ATC, SNOMED, and MeSH, as well as clinical guidelines and treatment protocols. 
It covers over 30 relationship types, including symptoms, treatments, and surgeries, with more than 1 million relationship instances.

\noindent \textbf {Evaluation Measures }
To establish consistent baseline results for future research, we employ several quantitative metrics for automatic evaluation. BLEU and ROUGE scores, well-established in NLP and multi-turn dialogue generation tasks, are used to assess the quality of generated responses by comparing them to ground truth~\cite{luo2018auto,li2017dailydialog}. 

\subsection{Dataset Statistics}
The basic statistics of the dataset are crucial for understanding its structure and quality. By analyzing the total number of dialogues, speaker turns, tokens, and key entities, we can evaluate the dataset's consistency and coverage. As shown in Table~\ref{static_result}, the dataset comprises 7,200 dialogues, with an average of 8.2 speaker turns and 131.1 tokens per dialogue, indicating a balanced and coherent dialogue structure. Moreover, the dataset includes 349,840 key entities, with an average of 48.6 key entities per dialogue, reflecting a high level of diversity and entity coverage. 
% These statistics offer valuable insights for training and evaluating models, ensuring they are well-suited for real-world dialogue scenarios.
\subsection{Automatic Evaluation}

We fine-tuned the selected Qwen2.5-14B-Instruct~\cite{qwen2.5} base model using the training set of our constructed multi-turn dialogue dataset with efficient LoRA parameter tuning. The fine-tuned model was then tested on the test set. We compared the test results with those of the LLaMA-3.1-8B-Instruct~\cite{dubey2024llama}, Qwen2.5-14B-Instruct, ChatGPT-3.5-Turbo, and ChatGPT-4o baselines. The automatic evaluation results are shown in Table ~\ref{automatic_result}. Fine-tuned Qwen2.5-14B-Instruct outperformed the other models across all evaluation metrics, particularly in BLEU-1 (0.391) and ROUGE-L (0.384), demonstrating its strong ability to generate text closely matching the reference. While ChatGPT-4o and Qwen2.5-14B-Instruct also performed well, they slightly lagged behind in higher-order BLEU scores. These results highlight the significant impact of fine-tuning on improving performance, especially for complex generation tasks.

\vspace{-1.5em}
\begin{table*}[ht]
% \small
\centering
\caption{\label{automatic_result}
Automatic Evaluation Results of Dialogue Generation Models.}
\resizebox{0.98\textwidth}{!}{
\renewcommand\arraystretch{0.8}

\begin{tabular}{llllll}
\toprule
                                   & BLEU-1   & BLEU-2    & BLEU-3 & BLEU-4& ROUGE-L  \\ \midrule
LLaMA-3.1-8B-Instruct           & 0.316   & 0.134   & 0.011  & 0.007 & 0.310 \\ \midrule
Qwen2.5-14B-Instruct  & 0.357  & 0.159   & 0.014 & 0.008&  0.344    \\ \midrule
ChatGPT-3.5-Turbo      & 0.347   & 0.142   & 0.014  & 0.008 &0.337  \\ \midrule
ChatGPT-4o         & 0.375   & 0.173   & 0.016  & 0.009 & 0.353  \\ \midrule
\bf{Fine-tuned Qwen2.5-14B-Instruct}   & \bf{0.391}   & \bf{0.188}   & \bf{0.022} & \bf{0.010} & \bf{0.384}    \\ \bottomrule
\end{tabular}
}
\vspace{-2.0em}
\end{table*}

\subsection{Human Evaluation} To assess the quality of our dataset, we conducted a manual evaluation by randomly selecting 200 dialogues from the training, validation, and test sets. Three domain experts rated the dialogues based on key metrics: coherence, question diversity and coverage, and semantic accuracy, with ratings ranging from 1 to 5. The results, shown in Table~\ref{human_result}, demonstrate strong performance across all metrics. High coherence scores indicate logical consistency, while the diversity and coverage scores reflect a broad range of question types. Strong semantic accuracy shows that answers are relevant and aligned with the questions. Overall, these findings confirm the dataset's quality, making it suitable for training and evaluating dialogue systems.

\vspace{-1.5em} % Add 
\begin{table}[ht!]
\centering
\caption{
Human Evaluation Results.}
\label{human_result}
% \resizebox{0.98\textwidth}{!}{
\tabcolsep=1.1em
\renewcommand\arraystretch{0.8}

\begin{tabular}{llll}
\toprule
                                   & train  & dev   & test  \\ \midrule
Coherence                    & 4.68   & 4.42   & 4.59    \\ \midrule
Diversity and Coverage & 4.26    & 4.37   & 4.45      \\ \midrule
Semantic Accurac        & 4.87  & 4.78 & 4.85   \\  
\bottomrule
\end{tabular}
% }
\vspace{-1.5em} % Add 
\end{table}

\subsection{Ablation Study: Effect of Community Partitioning and ARGW}

To evaluate the individual contributions of the Community Partitioning module and the Adaptive Relationship-guided GraphWalk (ARGW) algorithm, we conduct an ablation study with four settings:
First, the Baseline model generates dialogues without community partitioning or ARGW, performing random walks on the entire graph.
Second, Ablation 1 (No ARGW) applies community partitioning but performs random walks within each community.
Third, Ablation 2 (No Partitioning) uses ARGW on the full knowledge graph without partitioning.
Fourth, the Full Model integrates both community partitioning and ARGW.

We evaluate the models using BLEU and ROUGE-L. As shown in Table~\ref{tab:ablation}, the full model achieves the best performance, with BLEU-1 of 0.391 and ROUGE-L of 0.384. Removing either community partitioning or ARGW leads to performance drops, while the baseline performs the worst. These results confirm that both components are essential for generating high-quality, coherent dialogues.

\vspace{-1.5em}
\begin{table}[ht]
\centering
\caption{Ablation Study on the Impact of Community Partitioning and ARGW}
\label{tab:ablation}
\resizebox{0.98\textwidth}{!}{
\begin{tabular}{l|ccccc}
\hline
\textbf{Model Variant} & \textbf{BLEU-1} & \textbf{BLEU-2} & \textbf{BLEU-3} & \textbf{BLEU-4} & \textbf{ROUGE-L} \\
\hline
Baseline (No Partition, No ARGW)      & 0.305 & 0.121 & 0.010 & 0.006 & 0.298 \\
Ablation 1 (Partition Only)           & 0.332 & 0.138 & 0.012 & 0.007 & 0.319 \\
Ablation 2 (ARGW Only)                & 0.344 & 0.145 & 0.013 & 0.008 & 0.326 \\
Full Model (Partition + ARGW)         & \textbf{0.391} & \textbf{0.188} & \textbf{0.022} & \textbf{0.010} & \textbf{0.384} \\
\hline
\end{tabular}
}
\vspace{-1.5em}
\end{table}

%% file: main.bbl
\begin{thebibliography}{10}
\providecommand{\url}[1]{\texttt{#1}}
\providecommand{\urlprefix}{URL }
\providecommand{\doi}[1]{https://doi.org/#1}

\bibitem{abdullin2024synthetic}
Abdullin, Y., Molla-Aliod, D., Ofoghi, B., Yearwood, J., Li, Q.: Synthetic dialogue dataset generation using llm agents. arXiv preprint arXiv:2401.17461  (2024)

\bibitem{agarwal2020knowledge}
Agarwal, O., Ge, H., Shakeri, S., Al-Rfou, R.: Knowledge graph based synthetic corpus generation for knowledge-enhanced language model pre-training. arXiv preprint arXiv:2010.12688  (2020)

\bibitem{byambasuren2019preliminary}
Byambasuren, O., Yang, Y., Sui, Z., Dai, D., Chang, B., Li, S., Zan, H.: Preliminary study on the construction of chinese medical knowledge graph. Journal of Chinese Information Processing  \textbf{33}(10), ~1--9 (2019)

\bibitem{chen2020review}
Chen, X., Jia, S., Xiang, Y.: A review: Knowledge reasoning over knowledge graph. Expert systems with applications  \textbf{141},  112948 (2020)

\bibitem{christensen2024comparing}
Christensen, A.P., Garrido, L.E., Guerra-Pe{\~n}a, K., Golino, H.: Comparing community detection algorithms in psychometric networks: A monte carlo simulation. Behavior Research Methods  \textbf{56}(3),  1485--1505 (2024)

\bibitem{ding2023enhancing}
Ding, N., Chen, Y., Xu, B., Qin, Y., Zheng, Z., Hu, S., Liu, Z., Sun, M., Zhou, B.: Enhancing chat language models by scaling high-quality instructional conversations. arXiv preprint arXiv:2305.14233  (2023)

\bibitem{dubey2024llama}
Dubey, A., Jauhri, A., Pandey, A., Kadian, A., Al-Dahle, A., Letman, A., Mathur, A., Schelten, A., Yang, A., Fan, A., et~al.: The llama 3 herd of models. arXiv preprint arXiv:2407.21783  (2024)

\bibitem{feng-etal-2023-mmdialog}
Feng, J., Sun, Q., Xu, C., Zhao, P., Yang, Y., Tao, C., Zhao, D., Lin, Q.: {MMD}ialog: A large-scale multi-turn dialogue dataset towards multi-modal open-domain conversation. In: Rogers, A., Boyd-Graber, J., Okazaki, N. (eds.) Proceedings of the 61st Annual Meeting of the Association for Computational Linguistics (Volume 1: Long Papers). pp. 7348--7363. Association for Computational Linguistics, Toronto, Canada (Jul 2023)

\bibitem{kang2024artificial}
Kang, Y., Gao, S., Roth, R.E.: Artificial intelligence studies in cartography: a review and synthesis of methods, applications, and ethics. Cartography and Geographic Information Science pp. 1--32 (2024)

\bibitem{kumichev2024medsyn}
Kumichev, G., Blinov, P., Kuzkina, Y., Goncharov, V., Zubkova, G., Zenovkin, N., Goncharov, A., Savchenko, A.: Medsyn: Llm-based synthetic medical text generation framework. In: Joint European Conference on Machine Learning and Knowledge Discovery in Databases. pp. 215--230. Springer (2024)

\bibitem{lee2024llm2llm}
Lee, N., Wattanawong, T., Kim, S., Mangalam, K., Shen, S., Anumanchipalli, G., Mahoney, M.W., Keutzer, K., Gholami, A.: Llm2llm: Boosting llms with novel iterative data enhancement. arXiv preprint arXiv:2403.15042  (2024)

\bibitem{li2017dailydialog}
Li, Y., Su, H., Shen, X., Li, W., Cao, Z., Niu, S.: Dailydialog: A manually labelled multi-turn dialogue dataset. arXiv preprint arXiv:1710.03957  (2017)

\bibitem{li2023large}
Li, Y., Wang, S., Ding, H., Chen, H.: Large language models in finance: A survey. In: Proceedings of the fourth ACM international conference on AI in finance. pp. 374--382 (2023)

\bibitem{long2024llms}
Long, L., Wang, R., Xiao, R., Zhao, J., Ding, X., Chen, G., Wang, H.: On llms-driven synthetic data generation, curation, and evaluation: A survey. arXiv preprint arXiv:2406.15126  (2024)

\bibitem{luo2018auto}
Luo, L., Xu, J., Lin, J., Zeng, Q., Sun, X.: An auto-encoder matching model for learning utterance-level semantic dependency in dialogue generation. arXiv preprint arXiv:1808.08795  (2018)

\bibitem{mehta2024using}
Mehta, N., Goldwasser, D.: Using rl to identify divisive perspectives improves llms abilities to identify communities on social media. arXiv preprint arXiv:2406.00969  (2024)

\bibitem{meyer2023llm}
Meyer, L.P., Stadler, C., Frey, J., Radtke, N., Junghanns, K., Meissner, R., Dziwis, G., Bulert, K., Martin, M.: Llm-assisted knowledge graph engineering: Experiments with chatgpt. In: Working conference on Artificial Intelligence Development for a Resilient and Sustainable Tomorrow. pp. 103--115. Springer Fachmedien Wiesbaden Wiesbaden (2023)

\bibitem{construction-paired-knowledge-graph}
Mousavi, A., Zhan, X., Bai, R., Shi, P., Rekatsinas, T., Han, B., Li, Y., Pound, J., Susskind, J., Schluter, N., Ilyas, I., Jaitly, N.: Construction of paired knowledge graph - text datasets informed by cyclic evaluation. In: LREC-COLING (2024)

\bibitem{pan2024unifying}
Pan, S., Luo, L., Wang, Y., Chen, C., Wang, J., Wu, X.: Unifying large language models and knowledge graphs: A roadmap. IEEE Transactions on Knowledge and Data Engineering  (2024)

\bibitem{qiu2024smile}
Qiu, H., He, H., Zhang, S., Li, A., Lan, Z.: Smile: Single-turn to multi-turn inclusive language expansion via chatgpt for mental health support (2024)

\bibitem{stacey2024lucid}
Stacey, J., Cheng, J., Torr, J., Guigue, T., Driesen, J., Coca, A., Gaynor, M., Johannsen, A.: Lucid: Llm-generated utterances for complex and interesting dialogues (2024)

\bibitem{tang2023does}
Tang, R., Han, X., Jiang, X., Hu, X.: Does synthetic data generation of llms help clinical text mining? arXiv preprint arXiv:2303.04360  (2023)

\bibitem{qwen2.5}
Team, Q.: Qwen2.5: A party of foundation models (September 2024), \url{https://qwenlm.github.io/blog/qwen2.5/}

\bibitem{vuth-etal-2024-kgast}
Vuth, N., S{\'e}rasset, G., Schwab, D.: {KGAST}: From knowledge graphs to annotated synthetic texts. In: Biswas, R., Kaffee, L.A., Agarwal, O., Minervini, P., Singh, S., de~Melo, G. (eds.) Proceedings of the 1st Workshop on Knowledge Graphs and Large Language Models (KaLLM 2024). pp. 43--55. Association for Computational Linguistics, Bangkok, Thailand (Aug 2024)

\bibitem{wen2023re3dial}
Wen, J., Zhou, H., Guan, J., Huang, M.: Re$^3$dial: Retrieve, reorganize and rescale dialogue corpus for long-turn open-domain dialogue pre-training (2023)

\bibitem{wu2023bloomberggpt}
Wu, S., Irsoy, O., Lu, S., Dabravolski, V., Dredze, M., Gehrmann, S., Kambadur, P., Rosenberg, D., Mann, G.: Bloomberggpt: A large language model for finance. arXiv preprint arXiv:2303.17564  (2023)

\bibitem{xu2023knowledge}
Xu, R., Cui, H., Yu, Y., Kan, X., Shi, W., Zhuang, Y., Jin, W., Ho, J., Yang, C.: Knowledge-infused prompting: Assessing and advancing clinical text data generation with large language models  (2023)

\bibitem{yang2023refgpt}
Yang, D., Yuan, R., Fan, Y., Yang, Y., Wang, Z., Wang, S., Zhao, H.: Refgpt: Dialogue generation of gpt, by gpt, and for gpt (2023)

\bibitem{yang2023zhongjing}
Yang, S., Zhao, H., Zhu, S., Zhou, G., Xu, H., Jia, Y., Zan, H.: Zhongjing: Enhancing the chinese medical capabilities of large language model through expert feedback and real-world multi-turn dialogue (2023)

\bibitem{yi2024survey}
Yi, Z., Ouyang, J., Liu, Y., Liao, T., Xu, Z., Shen, Y.: A survey on recent advances in llm-based multi-turn dialogue systems. arXiv preprint arXiv:2402.18013  (2024)

\bibitem{zhao2023survey}
Zhao, W.X., Zhou, K., Li, J., Tang, T., Wang, X., Hou, Y., Min, Y., Zhang, B., Zhang, J., Dong, Z., et~al.: A survey of large language models. arXiv preprint arXiv:2303.18223  (2023)

\bibitem{zhou2024survey}
Zhou, Y., Guo, C., Wang, X., Chang, Y., Wu, Y.: A survey on data augmentation in large model era. arXiv preprint arXiv:2401.15422  (2024)

\bibitem{zou2020survey}
Zou, X.: A survey on application of knowledge graph. In: Journal of Physics: Conference Series. vol.~1487, p. 012016. IOP Publishing (2020)

\end{thebibliography}
